\documentclass{article}
\usepackage{spconf,amsmath}
\usepackage[final]{graphicx}
\usepackage{subcaption}
\usepackage{array}
\usepackage{tabu}
\usepackage{multirow}
\usepackage{makecell}

\usepackage{tikz}
\usetikzlibrary{shapes.geometric}
\usetikzlibrary{calc}
\newcommand{\toppage}{
  \begin{tikzpicture}[remember picture,overlay]
     \node[align=left, anchor=north west]
      at ($(current page.north west) + (1,-1)$)
      {978-1-5386-3640-4/19/\$31.00\textcopyright 2019 IEEE.\\Accepted for Publication at the IEEE International Symposium on Biomedical Imaging (ISBI 2019).};
  \end{tikzpicture}
}

\newcolumntype{M}[1]{>{\centering\arraybackslash}m{#1}}
\let\OLDthebibliography\thebibliography
\renewcommand\thebibliography[1]{
  \OLDthebibliography{#1}
  \setlength{\parskip}{2pt}
  \setlength{\itemsep}{0pt plus 0.3ex}
}

\title{DEEP CONVOLUTIONAL ENCODER-DECODERS WITH AGGREGATED MULTI-RESOLUTION SKIP CONNECTIONS FOR SKIN LESION SEGMENTATION}
\name{Author(s) Name(s)}
\address{Author Affiliation(s)}
\name{Ahmed H. Shahin$^{\star}$ \qquad Karim Amer$^{\dagger}$ \qquad Mustafa A. Elattar$^{\star}$}
\address{$^{\star}$ Medical Imaging and Image Processing Group, Center for Informatics Sciences, Nile University, Egypt \\
$^{\dagger}$ Ubiquitous and Visual Computing Group, Center for Informatics Sciences, Nile University, Egypt}
\begin{document}
\newcommand\addtag{\refstepcounter{equation}\tag{\theequation}}
%
\maketitle
\toppage{}
\begin{abstract}
The prevalence of skin melanoma is rapidly increasing as well as the recorded death cases of its patients. Automatic image segmentation tools play an important role in providing standardized computer-assisted analysis for skin melanoma patients. Current state-of-the-art segmentation methods are based on fully convolutional neural networks, which utilize an encoder-decoder approach. However, these methods produce coarse segmentation masks due to the loss of location information during the encoding layers. Inspired by Pyramid Scene Parsing Network (PSP-Net), we propose an encoder-decoder model that utilizes pyramid pooling modules in the deep skip connections which aggregate the global context and compensate for the lost spatial information. We trained and validated our approach using “ISIC 2018: Skin Lesion Analysis Towards Melanoma Detection” grand challenge dataset. Our approach showed a validation accuracy with a Jaccard index of 0.837, which outperforms U-Net. We believe that with this reported reliable accuracy, this method can be introduced for clinical practice.
\end{abstract}

\section{Introduction}
\label{sec:intro}

Skin cancer is by far the most common of all cancers. Skin melanoma incidents represent about 1\% of skin cancers but causes a large majority of skin cancer deaths. In 2018, it is estimated that more than 90,000 new incidents of melanoma will be diagnosed in the United States alone, resulting in 10,000 deaths \cite{Siegel2018Cancer2018}. Consequently, the early diagnosis of melanoma greatly increases the prevalence of recovery. The five-year survival rate for early stage melanoma exceeds 95\% \cite{SurvivalStage}. These statistics promote the importance of early diagnosis and identification of skin melanoma lesions. One of the essential steps in computerized analysis of dermoscopic images is automatic skin lesion segmentation.

A number of classical image processing techniques like thresholding, edge based, or region-based methods have been traditionally used for the sake of skin lesion segmentation \cite{Silveira2009ComparisonImages}. However, the existence of different sources of artifacts in images was a serious limitation to the classical techniques. For feature-based techniques, low-level and handcrafted features were not able to accurately segment skin lesions \cite{Hardie2018SkinFeatures}.

In the recent years, alongside with the advancements in the computational power of the graphical processing units (GPUs), Convolutional Neural Networks (CNNs) \cite{LeCun1998Gradient-basedRecognition} have emerged as one of the most powerful tools in image processing. CNN models have shown a promising performance in multiple domains including medical image analysis \cite{Ronneberger2015U-Net:Segmentation}. Long et al. \cite{Long2015FullySegmentation} introduced a fully convolutional network (FCN) for segmentation task where the input image is encoded using sequence of convolutions and pooling operators producing deep feature maps, and then decoded by a single deconvolutional upsampling layer to produce output feature maps with the original size. Despite the significant contribution over the classical machine learning and image processing methods, the FCN design had a negative effect on the output probability maps as the location information is distorted due to the sequential pooling operations producing coarse segmentation masks.

To alleviate the deficiencies of FCN, Badrinarayanan et al. \cite{Badrinarayanan2017SegNet:Segmentation} designed a trainable decoder with multiple deconvolutional layers operating on gradually upsampled feature maps. Ronneberger et al. \cite{Ronneberger2015U-Net:Segmentation} introduced skip connections to preserve important location information by concatenating the features from the contracting (encoding) path with the corresponding features in the expanding (decoding) path. PSP-Net \cite{Zhao2017PyramidNetwork} proposed a pyramid pooling module to aggregate the global context by using parallel pooling layers with differently sized kernels. Consequently, that provided additional contextual information preserved along the sequential convolution and pooling layers.

In skin lesion segmentation context, Yuan et al. \cite{Yuan2017ImprovingNetworks} utilized the conventional encoder-decoder architecture with a novel Jaccard-index-based loss function to handle the class imbalance in the dermoscopic images. Their FCN model has shown an improvement in the segmentation accuracy on ISIC 2016 dataset for skin lesion analysis \cite{Codella2018SkinISIC}. However, their study suffered from several limitations such as failing to achieve reasonable accuracy on some images that have low contrast between the lesion and the skin. Xue et al. \cite{Xue2018AdversarialSegmentation} explored the use of adversarial learning in the lesion segmentation task.

Inspired by PSP-Net, we propose an adjusted encoder-decoder architecture to overcome the coarse nature of the output segmentation map. Our main idea is to embed the pyramid pooling modules (PPMs) to the skip connections in the deeper convolutional layers. We show that such architecture aggregates the location information from the low convolutional layers, to alleviate the problem of spatial information loss. Our experiments on "ISIC 2018 Skin Lesion Analysis Towards Melanoma Detection" dataset \cite{Codella2018SkinISIC} corroborate our hypothesis that this approach refines the output predicted masks when compared to the current state-of-the-art lesion segmentation methods.
\begin{figure}[t]
\centering
\centerline{\includegraphics[height=5.7cm, width=8.4cm]{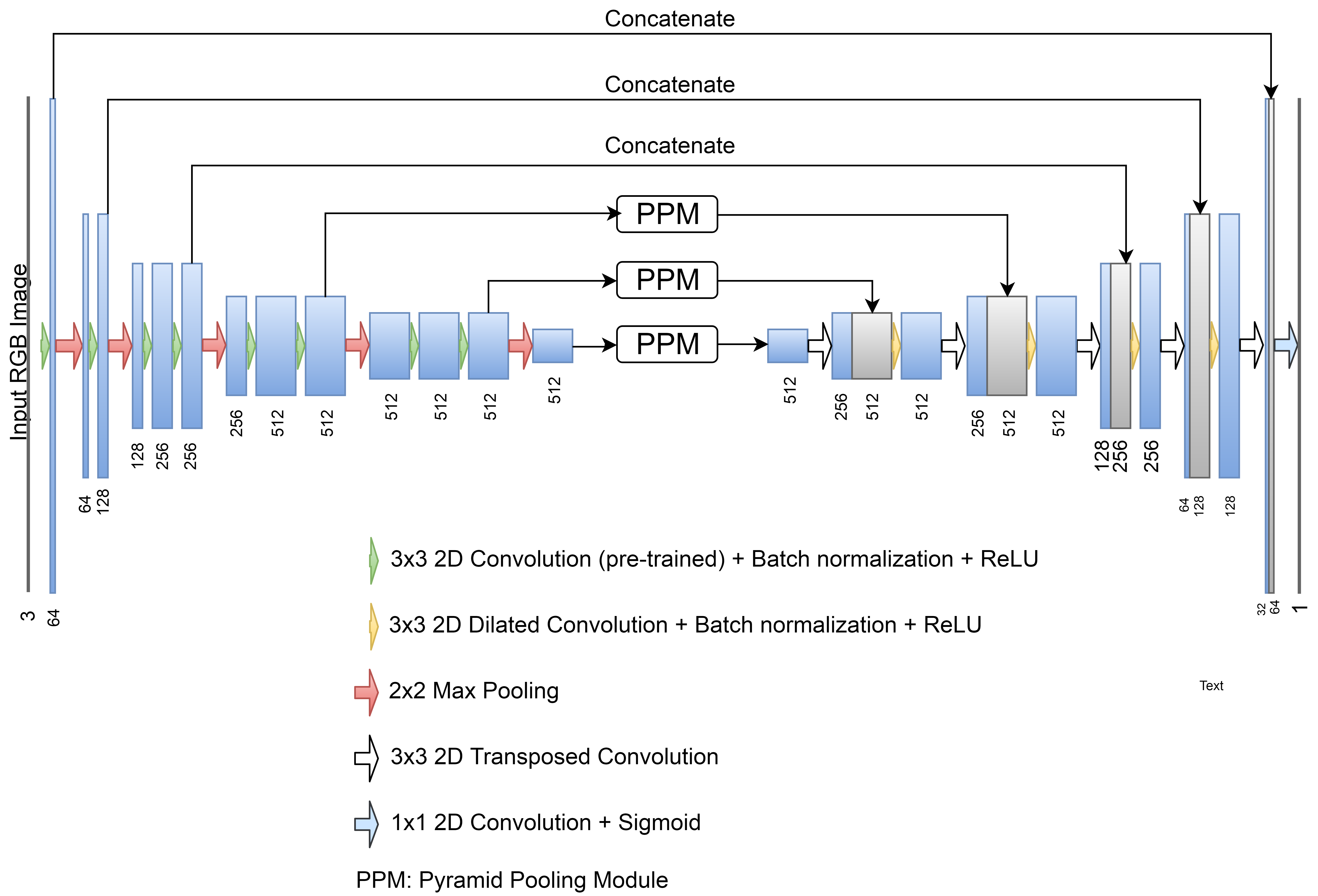}}
\caption{An illustration of the proposed architecture. Each blue box represents a multi-channel feature map. Number of channels is shown below each box. White boxes represent copied feature maps. The arrows denote the different operations. PPM: Pyramid Pooling Module.}
\end{figure}

\section{Methods}
\label{sec:pagestyle}

\textbf{Network Architecture}\hspace{4mm} Our proposed network architecture consists of an encoding path and a decoding path as shown in Fig. 1. For the encoding path, we use VGG11 \cite{Simonyan2014} network without the fully connected layers, with an additional batch normalization layer after every convolution. In the decoding path, the features map is upsampled using transposed convolutional layers which double the spatial resolution and reduces the number of channels by half. Additionally, each upsampled map is concatenated with the corresponding feature map from the encoding path through skip connections, then followed by a 3x3 dilated convolution, batch normalization, and ReLU. In all convolutional layers, we use padded convolutions in order to prevent the loss of borders.

As shown in Fig. 1, the two upper skip connections just concatenate the features from the encoding path with those in the decoding path to preserve the spatial information from the encoding path. However, the two lower skip connections projects the encoded activation maps to a Pyramid Pooling Module (PPM) [9] before concatenation. PPM covers different parts of the feature map through different pooling operations (see Fig. 2). An extra PPM block is used as a bottle neck between the encoding and the decoding path.

There is an output layer after the encoded path which performs a pixel-wise classification. The output layer is a 1x1 convolutional layer with a sigmoid activation function. The entire image is projected to a map of the same size, where each element represents the probability of its correspondence to the foreground (the lesion).

The output probability map is subjected to three successive post processing steps starting with thresholding the probability map at 0.5, then selecting the largest connected component and finally filling the holes of this component producing the final mask.
\begin{figure}[t]
\centering
\centerline{\includegraphics[height=4.6cm, width=5cm]{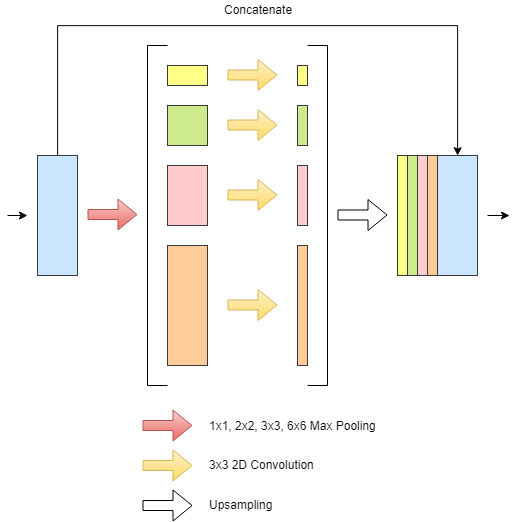}}
\caption{An illustration of the Pyramid Pooling Module (PPM).}
\end{figure}

\vspace{\baselineskip}
\noindent\textbf{Training}\hspace{4mm} Our model was implemented using PyTorch deep learning framework. Adam optimizer was used for neural network optimization. Dynamic learning rate was used and initialized by 5x10-5 then multiplied by 10-1 every 30 training epochs. The number of epochs was initially set to 200 epochs. However, in our experiments the network converged to the best validation accuracy in less than 100 epochs thanks to the pretrained weights. Additionally, the training batch size was set to 16 images. We augmented the training images using series of geometric transformations (horizontal flipping, vertical flipping, rotation, and zooming) to reduce model overfitting.
We used Generalized Dice Loss (GDL) \cite{Sudre2017GeneralisedSegmentations} as a loss function, which is a modified formula of dice score coefficient (DSC). Unlike DSC, GDL is differentiable and can be used as a loss function in case of imbalanced dataset, as an alternative for the widely used Cross-entropy loss. GDL takes the form:
\[ GDL = 1 - \frac{2 \sum_{n}^{}r_n p_n}{\sum_{n}{}r_n + \sum_{n}{}p_n} \addtag \]
Where: r is the reference ground truth segmentation with pixel values $r_n$, and p is the predicted probabilities from the CNN with pixel values $p_n$.
\section{Experiments and Results}
\label{sec:majhead}

Our data was extracted from the “ISIC 2018: Skin Lesion Analysis Towards Melanoma Detection” grand challenge datasets \cite{Codella2018SkinISIC}. The training set consists of 2594 RGB dermoscopic images with spatial resolutions ranging from 576 $\times$ 768 to 6748 $\times$ 4499. ISIC 2018 Validation and test data ground truth have not yet been released by the time of this paper submission. Thus, we divided the training data into 80\% training (2076 images) and 20\% validation set (518 images). The validation set was used to evaluate the performance of our approach. In order to guarantee the robustness of the model against different parts of the available data, five-fold cross-validation was used. We resized all the images to 192 $\times$ 256. However, the reported results were calculated after resampling the output masks to the original sizes. Fig. 3 shows the cross-validation results of our model. We compare our method with U-Net as a baseline against one random validation fold (518 images) of the dataset. Tables 1 \& 2 summarize the results of our method compared to U-Net, other approaches on ISIC 2017 dataset (2000 training images and 600 test images) and the results on the official test set (1000 images). We do not include results on ISIC 2016 dataset (900 training images and 375 test images) due to its relatively smaller number of test images. Using a Geforce 1080Ti Nvidia GPU, our method is able to segment around 10 images per second.
\begin{figure}[b]
\centering
\centerline{\includegraphics[height=4.8cm, width=6.5cm]{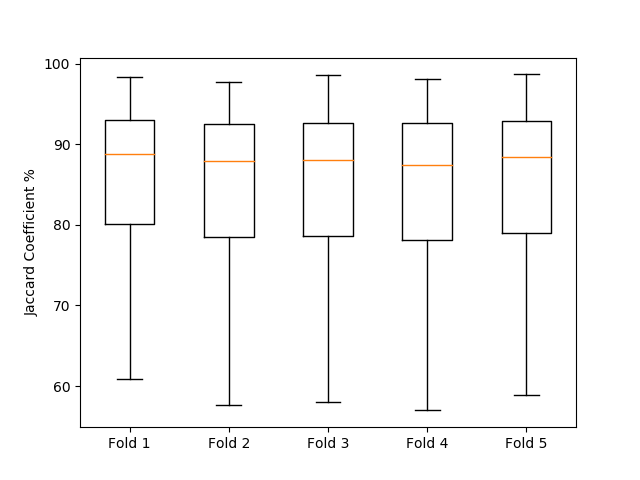}}
\caption{Five-fold cross-validation results of the proposed model on ISIC 2018 dataset.}
\end{figure}
\\
\begin{table}[h]
\caption {Results of our method against other methods evaluated on the local validation set. All values are in percentages. JA: Jaccard; DC: dice; SN: sensitivity; and SP: specificity.}
\label{tab:table_title}
\centering
\begin{tabular}{ | M{1.8cm} || M{1cm} | M{0.74cm} | M{0.74cm} | M{0.74cm} | M{0.74cm} | }
\hline
Method & Data & JA & DC & SN & SP\\
\hline
\end{tabular}
\begin{tabular}{ | M{1.8cm} || M{1cm} |M{0.74cm}| M{0.74cm} | M{0.74cm} | M{0.74cm} | }
\hline
Xue et al. \cite{Xue2018AdversarialSegmentation} & \multirow[c]{3}{=}{\makecell{ISIC \\ 2017}} & 78.5 & 86.7 & - & -\\
\cline{3-6} \cline{1-1}
Yuan et al. \cite{Yuan2017ImprovingNetworks} &  & 76.5 & 84.9 & 82.5 & \textbf{97.5}\\
\hline
U-Net \cite{Ronneberger2015U-Net:Segmentation} & \multirow{3}{=}{\makecell{ISIC \\ 2018}} & 82.6 & 89.5 & \textbf{91.1} & 96.8\\
\cline{3-6} \cline{1-1}
Proposed Method &  & \textbf{83.7} & \textbf{90.3} & 90.2 & 97.4\\
\hline
\end{tabular}
\centering
\end{table}
\begin{table}[h]
\caption {Results of our method against other methods evaluated on the official ISIC 2018 test set. The official evaluation metric was Jaccard thresholded at 65\%. All values are in percentages.}
\label{tab:title}
\centering
\begin{tabular}{ | M{3.5cm} || M{3.5cm} | }
\hline
Method & Thresholded Jaccard\\
\hline
\end{tabular}
\begin{tabular}{ | M{3.5cm} || M{3.5cm} | }
\hline
Bissoto et al. \cite{Bissoto2018Deep-Learning2018} & 72.8\\
\hline
Hardie et al. \cite{Hardie2018SkinFeatures} & 66.3\\
\hline
Proposed Method & \textbf{73.8}\\
\hline
\end{tabular}
\centering
\end{table}
\section{discussion}
\label{sssec:subsubhead}
Automatic lesion segmentation is still a challenging task due to the lack of distinctive lesion boundaries to adjacent skin, as well as the existence of various artifacts. In this work, we introduced an encoder-decoder like architecture equipped with PPMs for automatic skin lesion segmentation.
 
Our model has shown superior, and stable results on the five-fold cross-validation experiments. Compared to the previous approaches on smaller datasets, our method introduced a significant improvement in the segmentation accuracy. ISIC 2018 dataset is the latest publicly available dataset, that extends the previous ISIC 2017 and ISIC 2016 datasets with significantly more training and testing examples. Our model outperformed U-Net on our local validation set, we believe that this improvement is due to the PPMs role in preserving the spatial information along the encoding-decoding process by extracting additional contextual information in the deeper skip connections. Additionally, our method had a higher accuracy when compared to other published methods on the official test set of ISIC 2018 skin lesion analysis towards melanoma detection challenge. Bissoto et al. adopted the conventional U-Net architecture for their submission, while Hardie et al. used Bayesian classifier with handcrafted features and SVM regression for segmentation threshold selection. The difference between our scores on the local validation set and on the official test set is due to the 65\% thresholded Jaccard metric used in the test set evaluation. Our model does not rely on heavy augmentations or ensembling that would increase the prediction time. Fig. 4(a,b,c and d) shows success samples of the proposed method. There are entries with higher scores on the official leaderboard of ISIC challenge. However, there are no official presentations for their used methods.

Despite the superior performance in most of the cases, the model needs improvement to handle some failure cases. In Fig. 4(e,f), we can notice the difficulty of differentiating between the lesion and skin areas even for experienced raters, due to the very low contrast between the two classes. One way to further improve the segmentation performance is to use other post-processing techniques such as conditional random field and to combine its parameters in the network training process.
\begin{figure}[!t]
\centering
\begin{subfigure}{0.117\textwidth}
  \centering
  \includegraphics[width=0.95\linewidth]{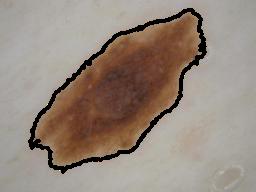}
  \caption{}
  \label{fig:sfig1}
\end{subfigure}%
\begin{subfigure}{0.117\textwidth}
  \centering
  \includegraphics[width=0.95\linewidth]{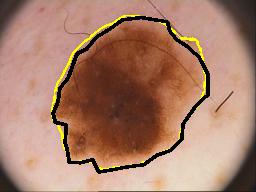}
  \caption{}
  \label{fig:sfig2}
\end{subfigure}
\begin{subfigure}{0.117\textwidth}
  \centering
  \includegraphics[width=0.95\linewidth]{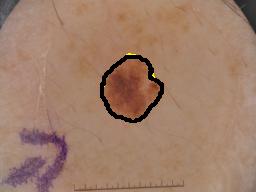}
  \caption{}
  \label{fig:sfig3}
\end{subfigure}
\\
\vspace{0.22cm}
\begin{subfigure}{0.114\textwidth}
  \centering
  \includegraphics[width=.95\linewidth]{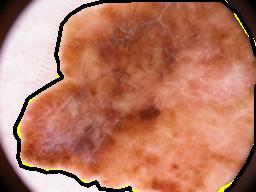}
  \caption{}
  \label{fig:sfig4}
\end{subfigure}
\begin{subfigure}{0.117\textwidth}
  \centering
  \includegraphics[width=0.95\linewidth]{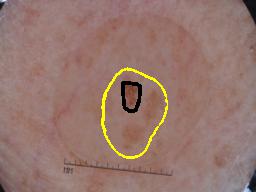}
  \caption{}
  \label{fig:sfig5}
\end{subfigure}
\begin{subfigure}{0.117\textwidth}
  \centering
  \includegraphics[width=.95\linewidth]{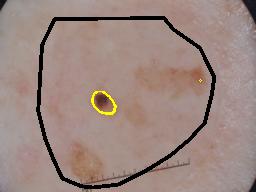}
  \caption{}
  \label{fig:sfig6}
\end{subfigure}
\caption{a, b, c and d: Success samples of our method. e and f: Failure samples of our method. Ground truth label is shown in black contour and the output predicted mask is in yellow.}
\label{fig:fig}
\end{figure}

\section{conclusions}
\label{sec:print}
In this work, we presented a fully automatic algorithm based on CNNs for skin lesion segmentation from dermoscopic images. We proposed the combination of encoder-decoder architectures with the pyramid pooling modules. Our method does not require any preprocessing or gray-scale conversion. According to the reported results, the method has shown its robustness against other methods and various image artifacts. We believe that this model can generalize well by showing reliable performance on other medical segmentation problems.

\vspace{\baselineskip}
\noindent\textbf{Acknowledgement}\hspace{4mm} In memory of Mohamed Samy Ahmed Kassem (September 1993 $-$ May 2018).
\bibliographystyle{Template_ISBI.bst}
\bibliography{Template_ISBI.bib}

\end{document}